\documentclass[conference]{IEEEtran}
\IEEEoverridecommandlockouts
\usepackage{cite}
\usepackage{amsmath,amssymb,amsfonts}
\usepackage{algorithmic}
\usepackage{graphicx}
\usepackage{textcomp}
\usepackage{xcolor}
\def\BibTeX{{\rm B\kern-.05em{\sc i\kern-.025em b}\kern-.08em
    T\kern-.1667em\lower.7ex\hbox{E}\kern-.125emX}}
\begin{document}

\title{Handwritten Digit Recognition using Machine and Deep Learning Algorithms}
\author{
\IEEEauthorblockN{\small Ritik Dixit}
\IEEEauthorblockA{\small \textit{Computer Science and Engineering} \\
\textit{{{\footnotesize Acropolis Institute of Technology \& Research}}}\\
\small Indore, India \\
\small dixitritik17@gmail.com}
\and
\IEEEauthorblockN{\small Rishika Kushwah}
\IEEEauthorblockA{\small \textit{Computer Science and Engineering} \\
\textit{{\footnotesize Acropolis Institute of Technology \& Research}}\\
\small Indore, India \\
\small rishikakushwah99@gmail.com}
\and
\IEEEauthorblockN{\small Samay Pashine}
\IEEEauthorblockA{\small \textit{Computer Science and Engineering} \\
\textit{{\footnotesize Acropolis Institute of Technology \& Research}}\\
\small Indore, India \\
\small samaypashine7@gmail.com
\and
\IEEEauthorblockN{\footnotesize}
\IEEEauthorblockA{\footnotesize} \\
\textit{{\footnotesize}}\\
\footnotesize \\
\footnotesize}
}

\maketitle

\begin{abstract}
The reliance of humans over machines has never been so high such that from object classification in photographs to adding sound to silent movies everything can be performed with the help of deep learning and machine learning algorithms. Likewise, Handwritten text recognition is one of the significant areas of research and development with a streaming number of possibilities that could be attained. Handwriting recognition (HWR), also known as Handwritten Text Recognition (HTR), is the ability of a computer to receive and interpret intelligible handwritten input from sources such as paper documents, photographs, touch-screens and other devices [1]. Apparently, in this paper, we have performed handwritten digit recognition with the help of MNIST datasets using Support Vector Machines (SVM), Multi-Layer Perceptron (MLP) and Convolution Neural Network (CNN) models. Our main objective is to compare the accuracy of the models stated above along with their execution time to get the best possible model for digit recognition. \\

\end{abstract}

\section{\textbf{INTRODUCTION}}
Handwritten digit recognition is the ability of a computer to recognize the human handwritten digits from different sources like images, papers, touch screens, etc, and classify them into 10 predefined classes (0-9). This has been a topic of boundless-research in the field of deep learning. Digit recognition has many applications like number plate recognition, postal mail sorting, bank check processing, etc [2]. In Handwritten digit recognition, we face many challenges because of different styles of writing of different peoples as it is not an Optical character recognition. 
This research provides a comprehensive comparison between different machine learning and deep learning algorithms for the purpose of handwritten digit recognition. For this, we have used Support Vector Machine, Multilayer Perceptron, and Convolutional Neural Network. The comparison between these algorithms is carried out on the basis of their accuracy, errors, and testing-training time corroborated by plots and charts that have been constructed using matplotlib for visualization.  \\

The accuracy of any model is paramount as more accurate models make better decisions. The models with low accuracy are not suitable for real-world applications. Ex- For an automated bank cheque processing system where the system recognizes the amount and date on the check, high accuracy is very critical. If the system incorrectly recognizes a digit, it can lead to major damage which is not desirable. That's why an algorithm with high accuracy is required in these real-world applications. Hence, we are providing a comparison of different algorithms based on their accuracy so that the most accurate algorithm with the least chances of errors can be employed in various applications of handwritten digit recognition. \\

This paper provides a reasonable understanding of machine learning and deep learning algorithms like SVM, CNN, and MLP for handwritten digit recognition. It furthermore gives you the information about which algorithm is efficient in performing the task of digit recognition. In further sections of this paper, we will be discussing the related work that has been done in this field followed by the methodology and implementation of all the three algorithms for the fairer understanding of them. Next, it presents the conclusion and result bolstered by the work we have done in this paper. Moreover, it will also give you some potential future enhancements that can be done in this field. The last section of this paper contains citations and references used. \\

\section{\textbf{RELATED WORK}}
With the humanization of machines, there has been a substantial amount of research and development work that has given a surge to deep learning and machine learning along with artificial intelligence. With time, machines are getting more and more sophisticated, from calculating the basic sums to doing retina recognition they have made our lives more secure and manageable. Likewise, handwritten text recognition is an important application of deep learning and machine learning which is helpful in detecting forgeries and a wide range of research has already been done that encompasses a comprehensive study and implementation of various popular algorithms like works done by S M Shamim [3], Anuj Dutt [4], Norhidayu binti [5] and Hongkai  Wang [8] to compare the different models of CNN with the fundamental machine learning algorithms on different grounds like performance rate, execution time, complexity and so on to assess each algorithm explicitly. [3] concluded that the Multilayer Perceptron classifier gave the most accurate results with minimum error rate followed by Support Vector Machine, Random Forest Algorithm, Bayes Net, Naïve Bayes, j48, and Random Tree respectively while [4] presented a comparison between SVM, CNN, KNN, RFC and were able to achieve the highest accuracy of 98.72\% using CNN (which took maximum execution time) and lowest accuracy using RFC. [5] did the detailed study-comparison on SVM, KNN and MLP models to classify the handwritten text and concluded that KNN and SVM predict all the classes of dataset correctly with 99.26\% accuracy but the thing process goes little complicated with MLP when it was having trouble classifying number 9, for which the authors suggested to use CNN with Keras to improve the classification. While [8] has focused on comparing deep learning methods with machine learning methods and comparing their characteristics to know which is better for classifying mediastinal lymph node metastasis of non-small cell lung cancer from 18 F-FDG PET/CT images and also to compare the discriminative power of the recently popular PET/CT texture features with the widely used diagnostic features. It concluded that the performance of CNN is not significantly different from the best classical methods and human doctors for classifying mediastinal lymph node metastasis of NSCLC from PET/CT images. However, CNN does not make use of the import diagnostic features, which have been proved more discriminative than the texture features for classifying small-sized lymph nodes. Therefore, incorporating the diagnostic features into CNN is a promising direction for future research. \\

All we need is lots of data and information and we will be able to train a big neural net to do what we want, so a convolution can be understood as "looking at functions surrounding to make a precise prognosis of its outcome." [6], [7] has used a convolution neural network for handwritten digit recognition using MNIST datasets. [6] has used 7 layered CNN model with 5 hidden layers along with gradient descent and back prorogation model to find and compare the accuracy on different epochs, thereby getting maximum accuracy of 99.2\% while in [7], they have briefly discussed different components of CNN, its advancement from LeNet-5 to SENet and comparisons between different model like AlexNet, DenseNet and ResNet. The research outputs the LeNet-5 and LeNet-5 (with distortion) achieved test error rate of 0.95\% and 0.8\% respectively on MNIST data set, the architecture and accuracy rate of AlexNet is same as LeNet-5 but much bigger with around 4096000 parameters and ”Squeeze-and-Excitation network” (SENet) have become the winner of ILSVRC-2017 since they have reduced the top-5 error rate to 2.25\% and by far the most sophisticated model of CNN in existence.\\

\section{\textbf{METHODOLOGY}}
The comparison of the algorithms (Support vector machines, Multi-layered perceptron and Convolutional neural network) is based on the characteristic chart of each algorithm on common grounds like dataset, the number of epochs, complexity of the algorithm, accuracy of each algorithm, specification of the device (Ubuntu 20.04 LTS, i5 7th gen processor) used to execute the program and runtime of the algorithm, under ideal condition. \\

\begin{flushleft}
\textbf{\textit{A. 		DATASET}}
\end{flushleft}
Handwritten character recognition is an expansive research area that already contains detailed ways of implementation which include major learning datasets, popular algorithms, features scaling and feature extraction methods. MNIST dataset (Modified National Institute of Standards and Technology database) is the subset of the NIST dataset which is a combination of two of NIST's databases: Special Database 1 and Special Database 3. Special Database 1 and Special Database 3 consist of digits written by high school students and employees of the United States Census Bureau, respectively. MNIST contains a total of 70,000 handwritten digit images (60,000 - training set and 10,000 - test set) in 28x28 pixel bounding box and anti-aliased. All these images have corresponding Y values which apprises what the digit is. \\

\begin{center}
\includegraphics[scale=0.65]{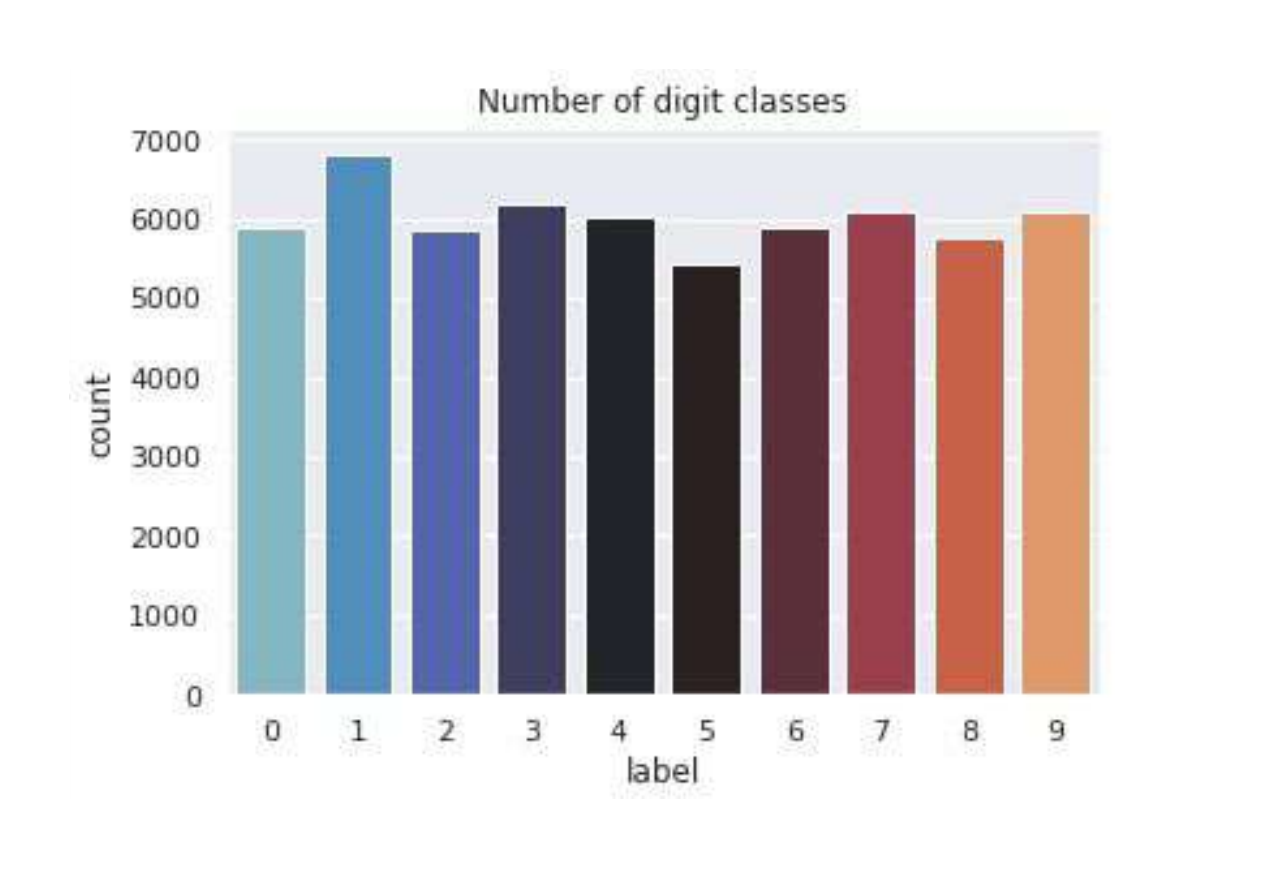}
\end{center}
\footnotesize Figure 1. Bar graph illustrating the MNIST handwritten digit training dataset (Label vs Total number of training samples). \\

\begin{center}
\includegraphics[scale=0.55]{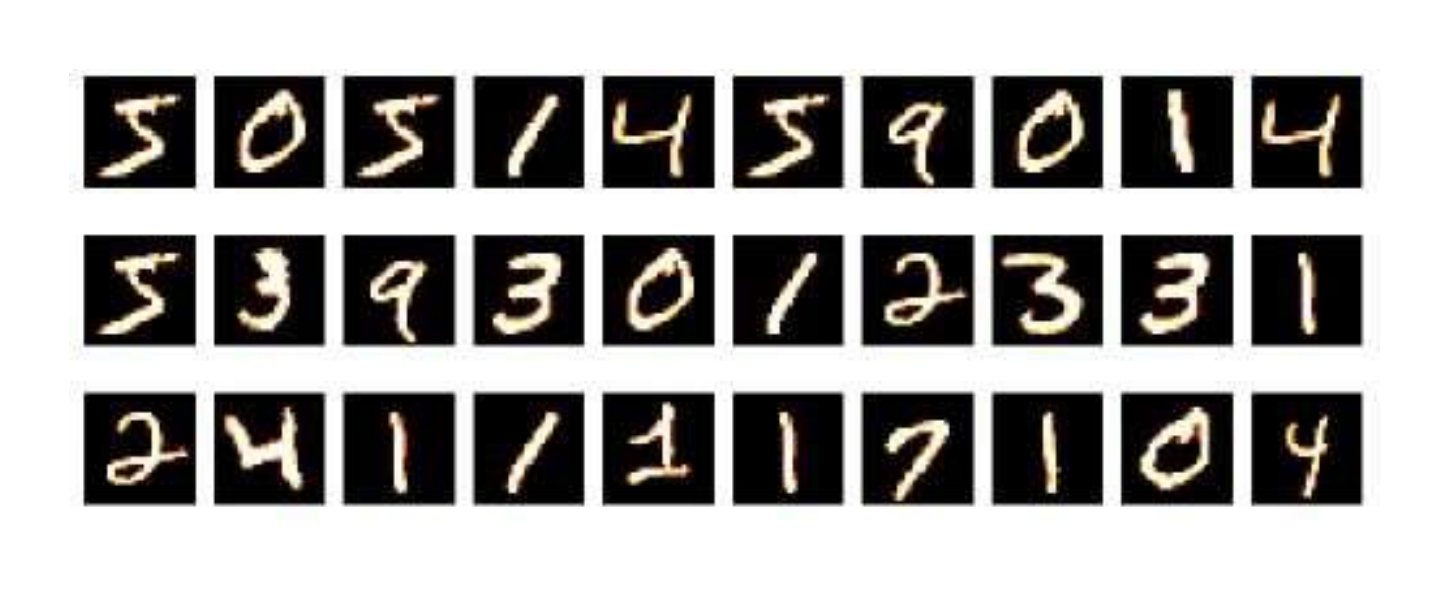}
\end{center}
\footnotesize Figure 2. Plotting of some random MNIST Handwritten digits.
\\
\begin{flushleft}
\textbf{\textit{B.		SUPPORT VECTOR MACHINE}}
\end{flushleft}
Support Vector Machine (SVM) is a supervised machine learning algorithm. In this, we generally plot data items in n-dimensional space where n is the number of features, a particular coordinate represents the value of a feature, we perform the classification by finding the hyperplane that distinguishes the two classes. It will choose the hyperplane that separates the classes correctly. SVM chooses the extreme vectors that help in creating the hyperplane. These extreme cases are called support vectors, and hence the algorithm is termed as Support Vector Machine. There are mainly two types of SVMs, linear and non-linear SVM. In this paper, we have used Linear SVM for handwritten digit recognition [10].

\begin{center}
\includegraphics[scale=0.55]{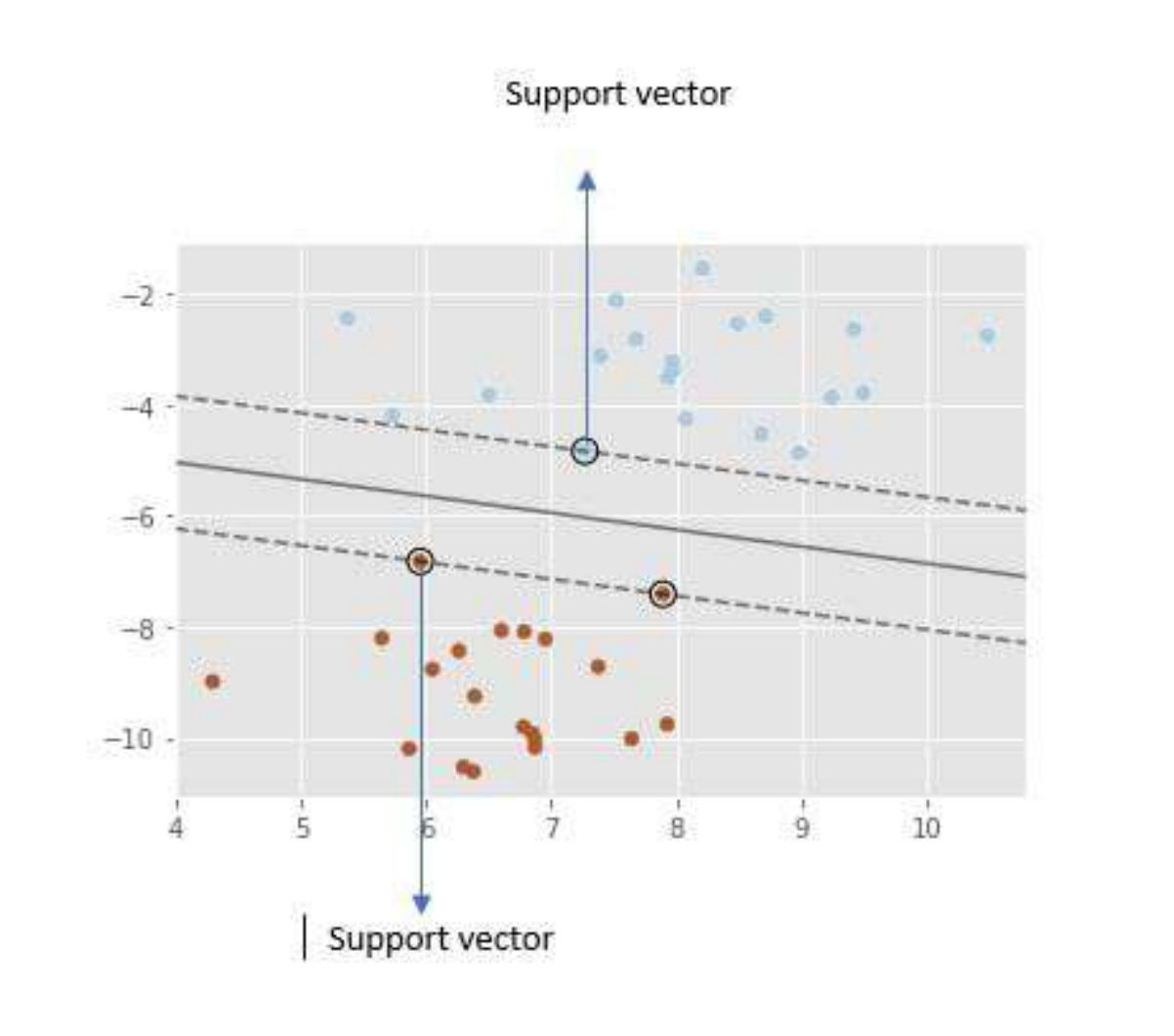}
\end{center}
\footnotesize Figure 3. This image describes the working mechanism of SVM Classification with supporting vectors and hyperplanes. \\

\begin{flushleft}
\textbf{\textit{C.		MULTILAYERED PERCEPTRON}}
\end{flushleft}
A multilayer perceptron (MLP) is a class of feedforward artificial neural networks (ANN). It consists of three layers: input layer, hidden layer and output layer. Each layer consists of several nodes that are also formally referred to as neurons and each node is interconnected to every other node of the next layer. In basic MLP there are 3 layers but the number of hidden layers can increase to any number as per the problem with no restriction on the number of nodes. The number of nodes in the input and output layer depends on the number of attributes and apparent classes in the dataset respectively. The particular number of hidden layers or numbers of nodes in the hidden layer is difficult to determine due to the model erratic nature and therefore selected experimentally. Every hidden layer of the model can have different activation functions for processing. For learning purposes, it uses a supervised learning technique called backpropagation. In the MLP, the connection of the nodes consists of a weight that gets adjusted to synchronize with each connection in the training process of the model[11].

\begin{center}
\includegraphics[scale=0.65]{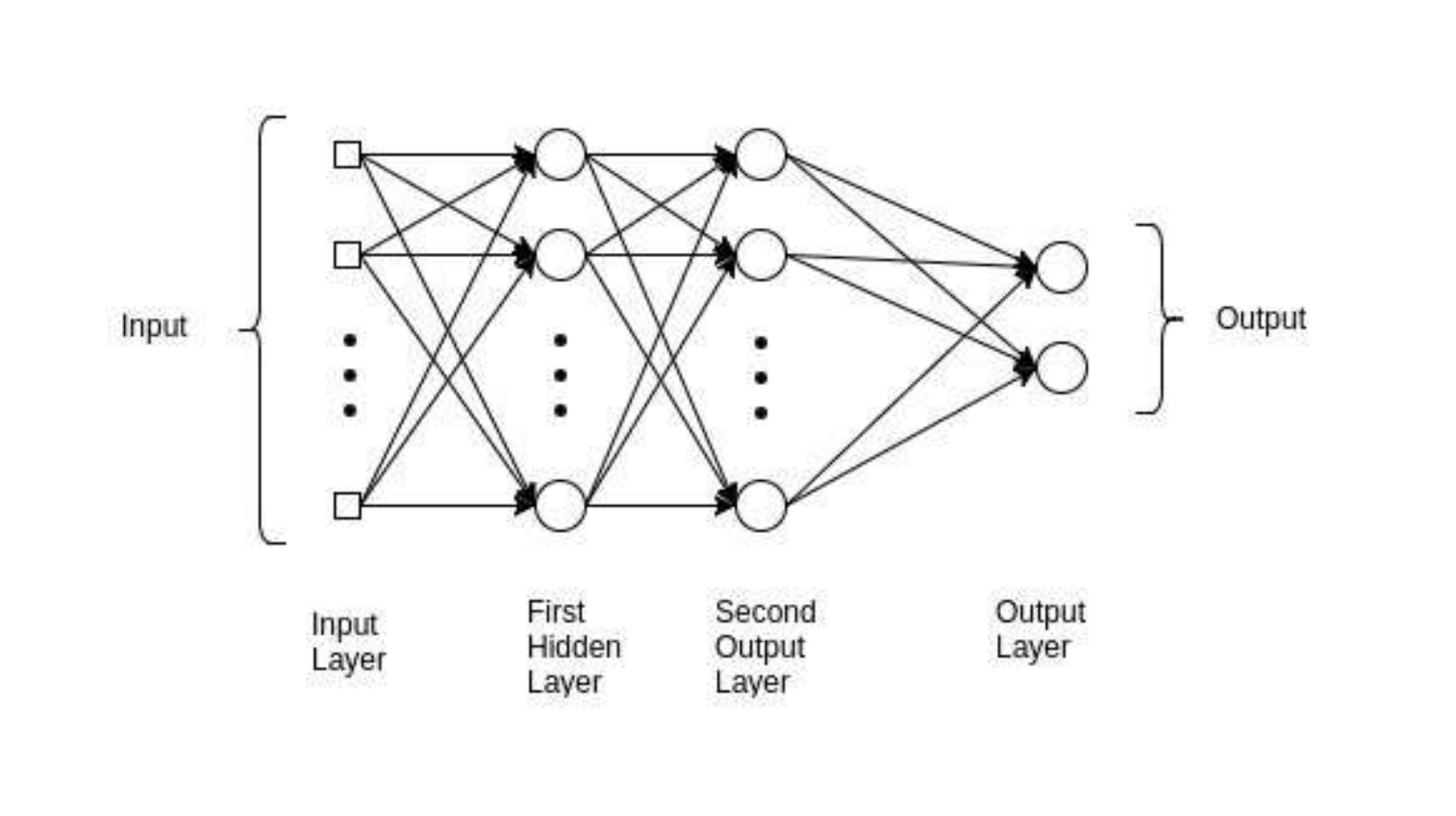}
\end{center}
\footnotesize Figure 4. This figure illustrates the basic architecture of the Multilayer perceptron with variable specification of the network.

\begin{flushleft}
\textbf{\textit{D.		CONVOLUTIONAL NEURAL NETWORK}}
\end{flushleft}
CNN is a deep learning algorithm that is widely used for image recognition and classification. It is a class of deep neural networks that require minimum pre-processing. It inputs the image in the form of small chunks rather than inputting a single pixel at a time, so the network can detect uncertain patterns (edges) in the image more efficiently. CNN contains 3 layers namely, an input layer, an output layer, and multiple hidden layers which include Convolutional layers, Pooling layers(Max and Average pooling), Fully connected layers (FC), and normalization layers [12]. CNN uses a filter (kernel) which is an array of weights to extract features from the input image. CNN employs different activation functions at each layer to add some non-linearity [13]. As we move into the CNN, we observe the height and width decrease while the number of channels increases. Finally, the generated column matrix is used to predict the output [14]. \\

\begin{center}
\includegraphics[scale=0.37]{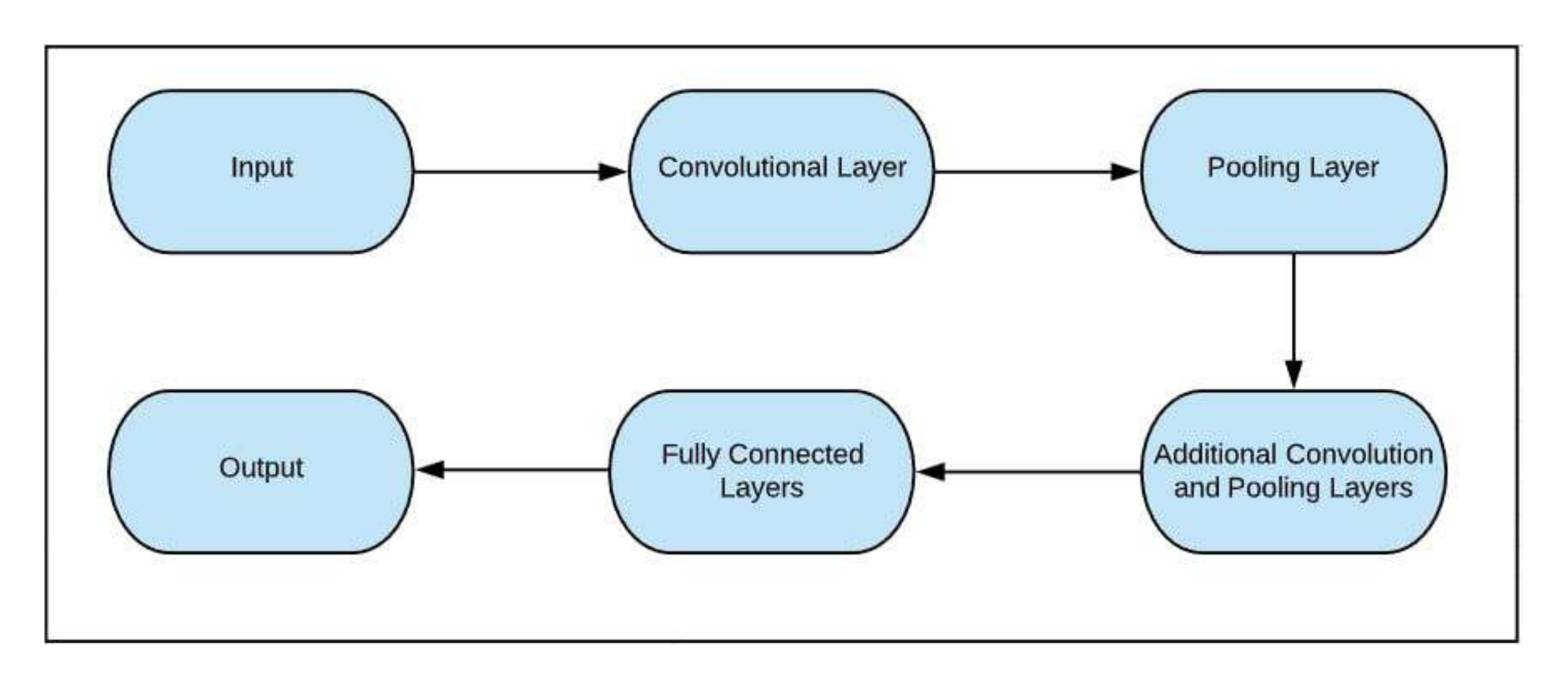}
\end{center}
\footnotesize Figure 5. This figure shows the architectural design of CNN layers in the form of a Flow chart. \\

\begin{flushleft}
\textbf{\textit{E.		VISUALIZATION}}
\end{flushleft}
In this research, we have used the MNIST dataset (i.e. handwritten digit dataset) to compare different level algorithm of deep and machine learning (i.e. SVM, ANN-MLP, CNN) on the basis of execution time, complexity, accuracy rate, number of epochs and number of hidden layers (in the case of deep learning algorithms). To visualize the information obtained by the detailed analysis of algorithms we have used bar graphs and tabular format charts using module matplotlib, which gives us the most precise visuals of the step by step advances of the algorithms in recognizing the digit. The graphs are given at each vital part of the programs to give visuals of each part to bolster the outcome. \\

\section{\textbf{IMPLEMENTATION}}
To compare the algorithms based on working accuracy, execution time, complexity, and the number of epochs (in deep learning algorithms) we have used three different classifiers: \\
\begin{itemize}
	\item Support Vector Machine Classifier 
	\item ANN - Multilayer Perceptron Classifier 
	\item Convolutional Neural Network Classifier 
\end{itemize}

We have discussed in detail about the implementation of each algorithm explicitly below to create a flow of this analysis to create a fluent and accurate comparison. \\

\begin{flushleft}
\textbf{\textit{I. 		PRE-PROCESSING}}
\end{flushleft}
Pre-processing is an initial step in the machine and deep learning which focuses on improving the input data by reducing unwanted impurities and redundancy. To simplify and break down the input data we reshaped all the images present in the dataset in 2-dimensional images i.e (28,28,1). Each pixel value of the images lies between 0 to 255 so, we Normalized these pixel values by converting the dataset into 'float32' and then dividing by 255.0 so that the input features will range between 0.0 to 1.0. Next, we performed one-hot encoding to convert the y values into zeros and ones, making each number categorical, for example, an output value 4 will be converted into an array of zero and one i.e [0,0,0,0,1,0,0,0,0,0]. \\

\begin{flushleft}
\textbf{\textit{II. 	SUPPORT VECTOR MACHINE}}
\end{flushleft}
The SVM in scikit-learn [16] supports both dense (numpy.ndarray and convertible to that by numpy.asarray) and sparse (any scipy.sparse) sample vectors as input. In scikit-learn, SVC, NuSVC and LinearSVC are classes capable of performing multi-class classification on a dataset. In this paper we have used LinearSVC for classification of MNIST datasets that make use of a Linear kernel implemented with the help of LIBLINEAR [17]. \\

Various scikit-learn libraries like NumPy, matplotlib, pandas, Sklearn and seaborn have been used for the implementation purpose. Firstly, we will download the MNIST datasets, followed by loading it and reading those CSV files using pandas. \\

After this, plotting of some samples as well as converting into matrix followed by normalization and scaling of features have been done. Finally, we have created a linear SVM model and confusion matrix that is used to measure the accuracy of the model [9]. \\

\begin{flushleft}
\textbf{\textit{III. 	MULTILAYERED PERCEPTRON}}
\end{flushleft}
The implementation of Handwritten digits recognition by Multilayer perceptron [18] which is also known as feedforward artificial neural network is done with the help of Keras module to create an MLP model of Sequential class and add respective hidden layers with different activation function to take an image of 28x28 pixel size as input. After creating a sequential model, we added a Dense layer of different specifications and Drop out layers as shown in the image below. The block diagram is given here for reference. Once you have the training and test data, you can follow these steps to train a neural network in Keras. \\

\begin{flushleft}
\includegraphics[scale=0.35]{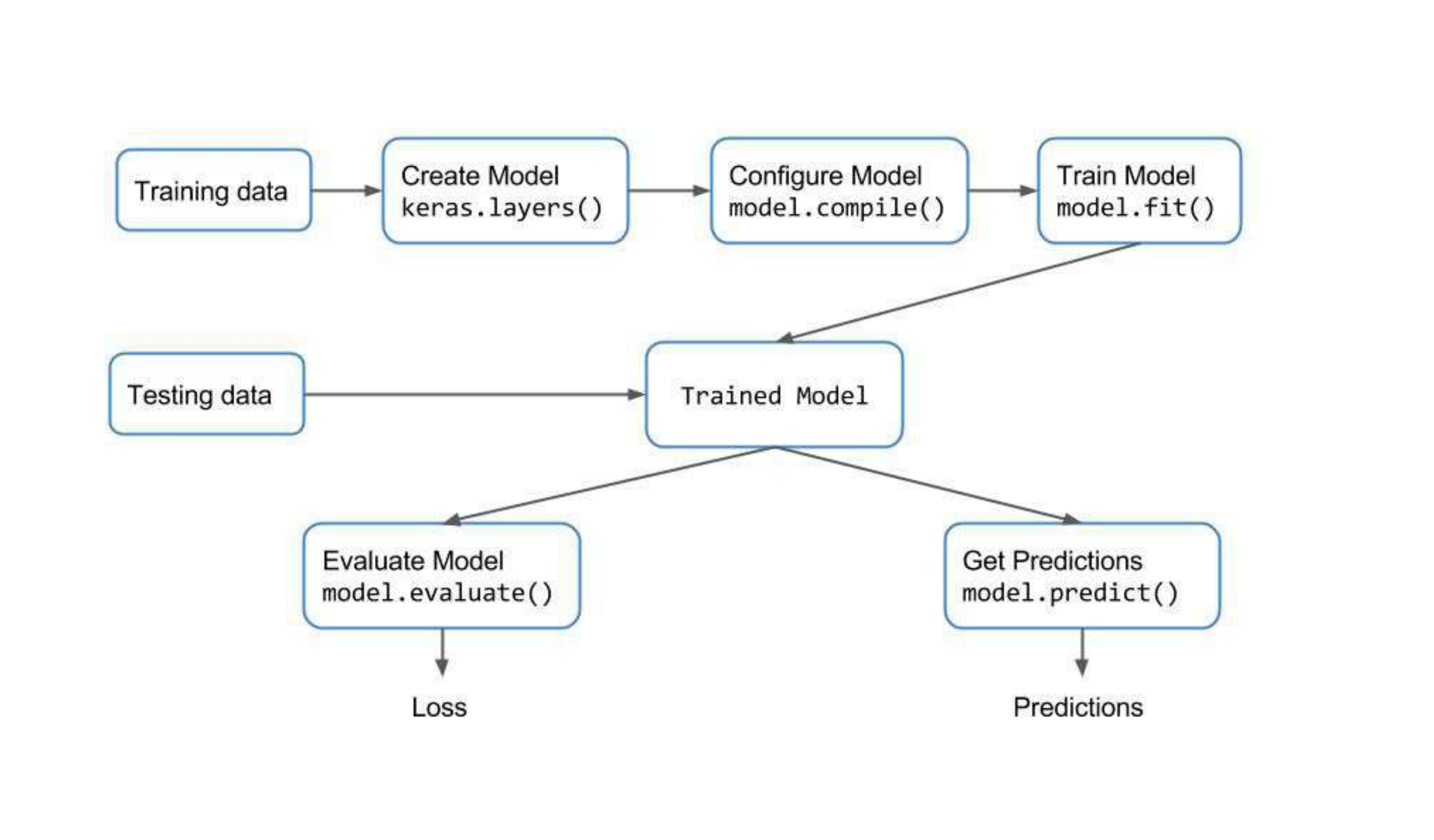}
\end{flushleft}
\footnotesize Figure 6.  Sequential Block Diagram of Multi-layers perceptron model built with the help of Keras module [19]. \\

We used a neural network with 4 hidden layers and an output layer with 10 units (i.e. total number of labels). The number of units in the hidden layers is kept to be 512. The input to the network is the 784-dimensional array converted from the 28×28 image. We used the Sequential model for building the network. In the Sequential model, we can just stack up layers by adding the desired layer one by one. We used the Dense layer, also called a fully connected layer since we are building a feedforward network in which all the neurons from one layer are connected to the neurons in the previous layer. Apart from the Dense layer, we added the ReLU activation function which is required to introduce non-linearity to the model. This will help the network learn non-linear decision boundaries. The last layer is a softmax layer as it is a multiclass classification problem [19].\\

\begin{flushleft}
\textbf{\textit{IV. 	CONVOLUTIONAL NEURAL NETWORK}}
\end{flushleft}
The implementation of handwritten digit recognition by Convolutional Neural Network [15] is done using Keras. It is an open-source neural network library that is used to design and implement deep learning models. From Keras, we have used a Sequential class which allowed us to create model layer-by-layer. The dimension of the input image is set to 28(Height), 28(Width), 1(Number of channels). Next, we created the model whose first layer is a Conv layer [20]. This layer uses a matrix to convolve around the input data across its height and width and extract features from it. This matrix is called a Filter or Kernel. The values in the filter matrix are weights. We have used 32 filters each of the dimensions (3,3) with a stride of 1. Stride determines the number of pixels shifts. Convolution of filter over the input data gives us activation maps whose dimension is given by the formula: ((N + 2P - F)/S) + 1 where N= dimension of input image, P= padding, F= filter dimension and S=stride. In this layer, Depth (number of channels) of the output image is equal to the number of filters used. To increase the non-linearity, we have used an activation function that is Relu [21]. Next, another convolutional layer is used in which we have applied 64 filters of the same dimensions (3,3) with a stride of 1 and the Relu function. \\

Next, to these layers, the pooling layer [22] is used which reduces the dimensionality of the image and computation in the network. We have employed MAX-pooling which keeps only the maximum value from a pool. The depth of the network remains unchanged in this layer. We have kept the pool-size (2,2) with a stride of 2, so every 4 pixels will become a single pixel. To avoid overfitting in the model, Dropout layer [23] is used which drops some neurons which are chosen randomly so that the model can be simplified. We have set the probability of a node getting dropped out to 0.25 or 25\%. Following it, Flatten Layer [23] is used which involves flattening i.e. generating a column matrix (vector) from the 2-dimensional matrix. This column vector will be fed into the fully connected layer [24]. This layer consists of 128 neurons with a dropout probability of 0.5 or 50\%. After applying the Relu activation function, the output is fed into the last layer of the model that is the output layer. This layer has 10 neurons that represent classes (numbers from 0 to 9) and the SoftMax function [25] is employed to perform the classification. This function returns probability distribution over all the 10 classes. The class with the maximum probability is the output. \\

\begin{flushleft}
\includegraphics[scale=0.175]{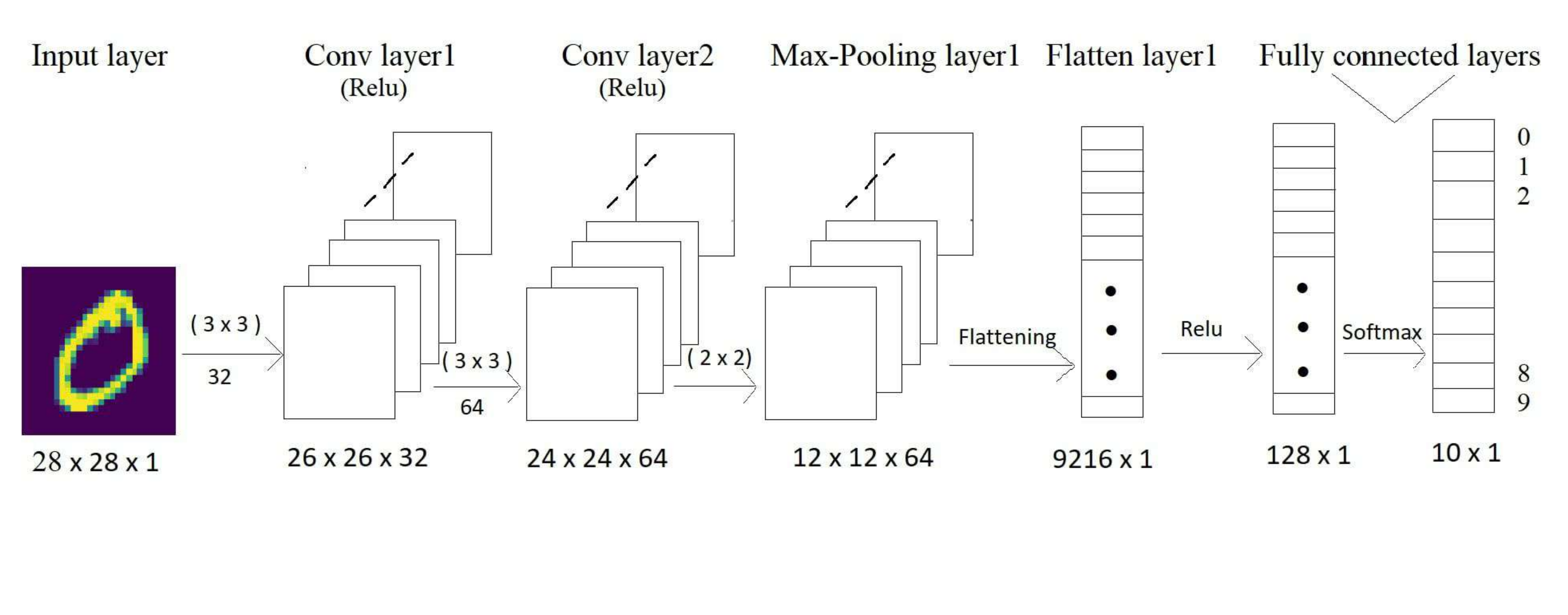}
\end{flushleft}
\footnotesize Figure 7. Detailed architecture of Convolutional Neural Network with apt specifications of each layer
\section{\textbf{RESULT}}
\begin{small}
After implementing all the three algorithms that are SVM, MLP and CNN we have compared their accuracies and execution time with the help of experimental graphs for perspicuous understanding. We have taken into account the Training and Testing Accuracy of all the models stated above. After executing all the models, we found that SVM has the highest accuracy on training data while on testing dataset CNN accomplishes the utmost accuracy. Additionally, we have compared the execution time to gain more insight into the working of the algorithms. Generally, the running time of an algorithm depends on the number of operations it has performed. So, we have trained our deep learning model up to 30 epochs and SVM models according to norms to get the apt outcome. SVM took the minimum time for execution while CNN accounts for the maximum running time. 

\begin{footnotesize}
\begin{table}[h!]
  \begin{center}
    \caption{Comparison Analysis of different models.}
    \label{tab:table1}
    \begin{tabular}{c|c|c|c} 
      \textbf{MODEL} & \textbf{TRAINING RATE} & \textbf{TESTING RATE} & \textbf{EXECUTION TIME}\\
      \hline
      SVM & 99.98\% & 94.005\% & 1:35 min \\
      MLP & 99.92\% & 98.85\% & 2:32 min \\
      CNN & 99.53\% & 99.31\% & 44:02 min \\
    \end{tabular}
  \end{center}
\end{table}
\end{footnotesize}
\footnotesize Figure 8. This table represents the overall performance for each model. The table contains 5 columns, the 2nd column represents model name, 3rd and 4th column represents the training and testing accuracy of models, and 5th column represents execution time of models.

\begin{center}
\includegraphics[scale=0.3]{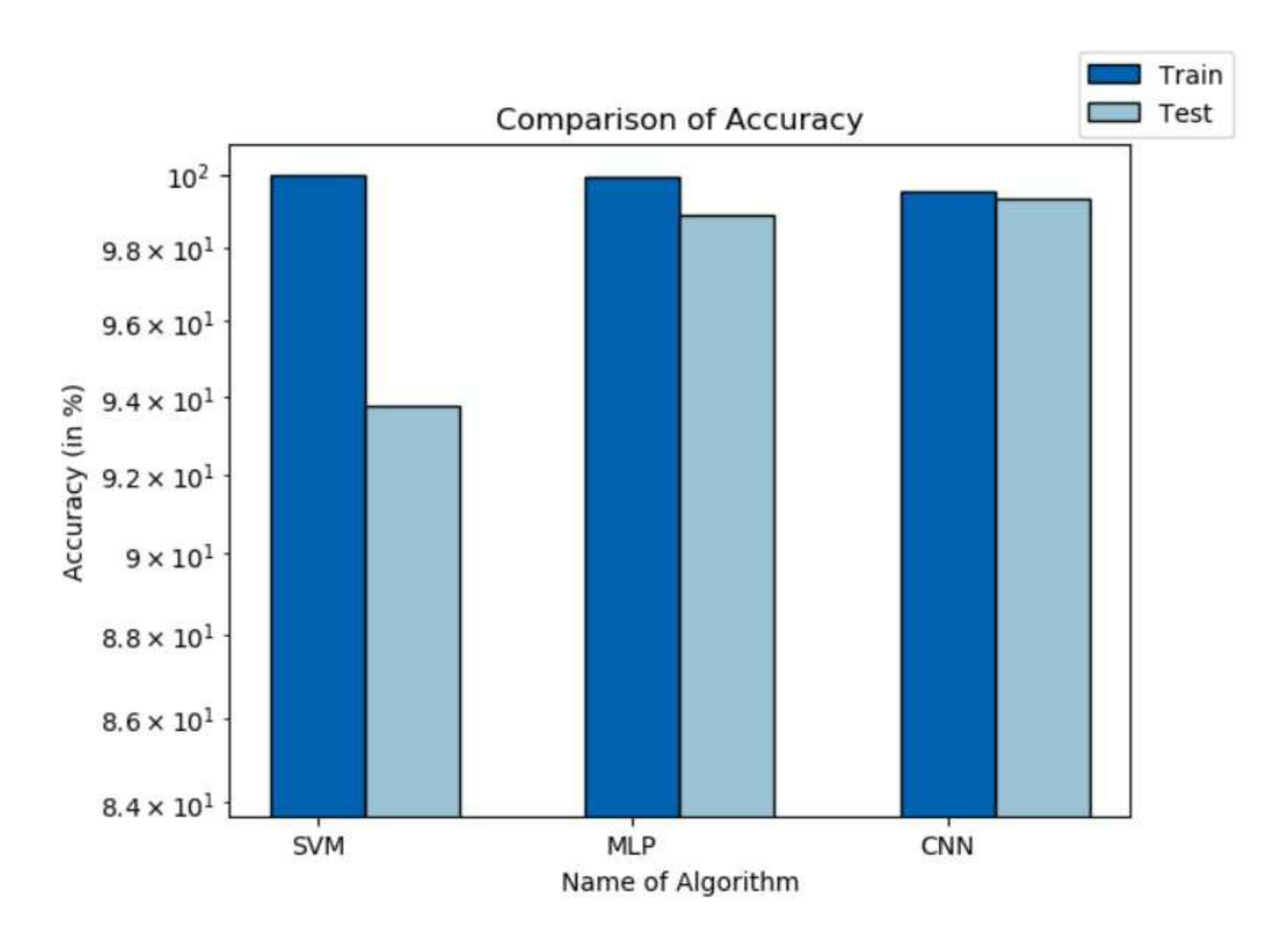}
\end{center}
\footnotesize Figure 9. Bar graph depicting accuracy comparison  (SVM (Train: 99.98\%, Test: 93.77\%), MLP (Train: 99.92\%, Test: 98.85\%), CNN (Train: 99.53\%, Test: 99.31\%)). 

\begin{center}
\includegraphics[scale=0.33]{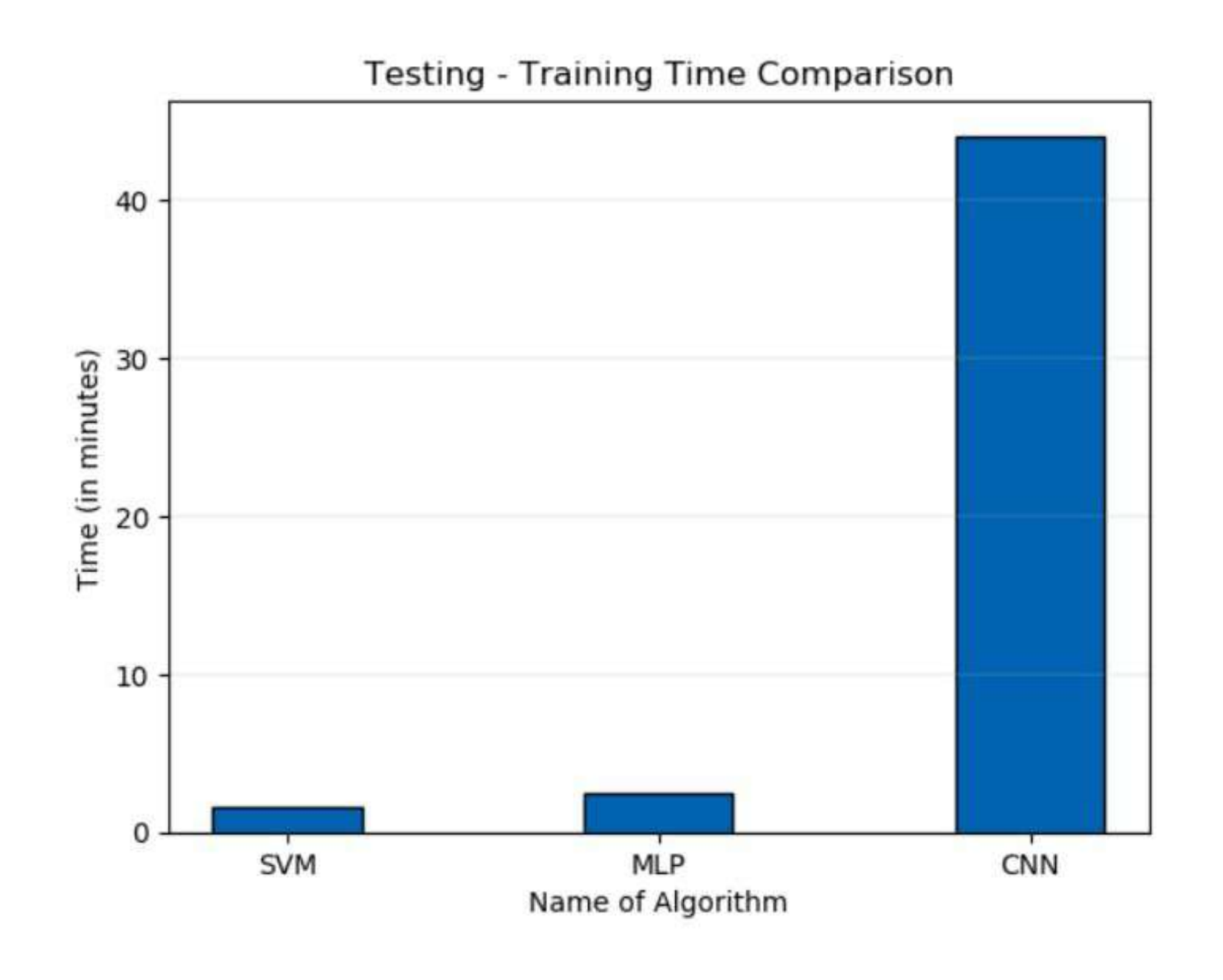}
\end{center}
\footnotesize Figure 10. Bar graph showing execution time comparison of SVM, MLP and CNN                       (SVM: 1.58 mins, MLP: 2.53 mins, CNN: 44.05 mins).\\

Furthermore, we visualized the performance measure of deep learning models and how they ameliorated their accuracy and reduced the error rate concerning the number of epochs. The significance of sketching the graph is to know where we should apply early stop so that we can avoid the problem of overfitting as after some epochs, change in accuracy becomes constant.

\begin{center}
\includegraphics[scale=0.5]{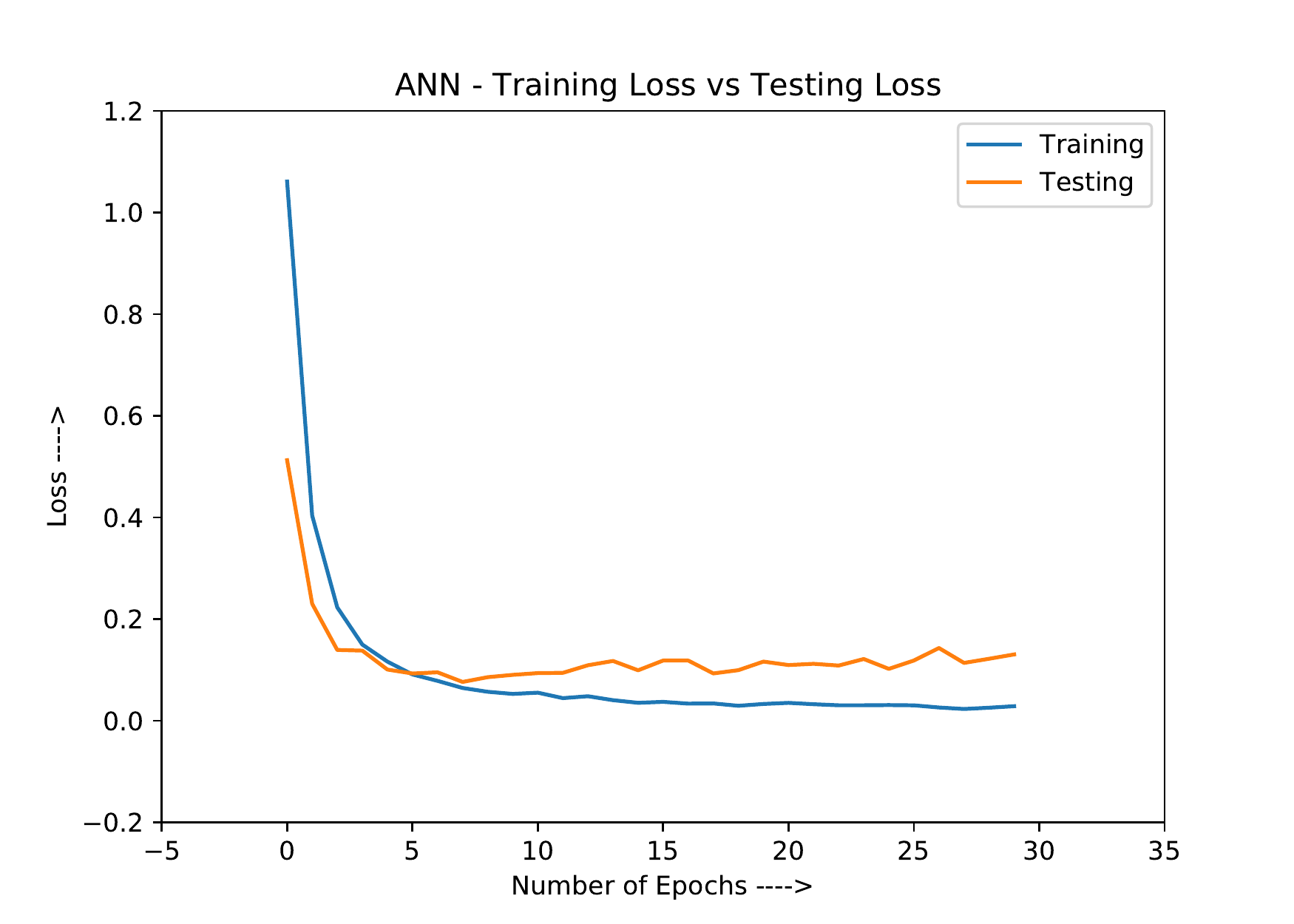}
\end{center}
\footnotesize Figure 11. Graph illustrating the transition of training loss with increasing number of epochs in Multilayer Perceptron (Loss rate v/s Number of epochs).

\begin{center}
\includegraphics[scale=0.5]{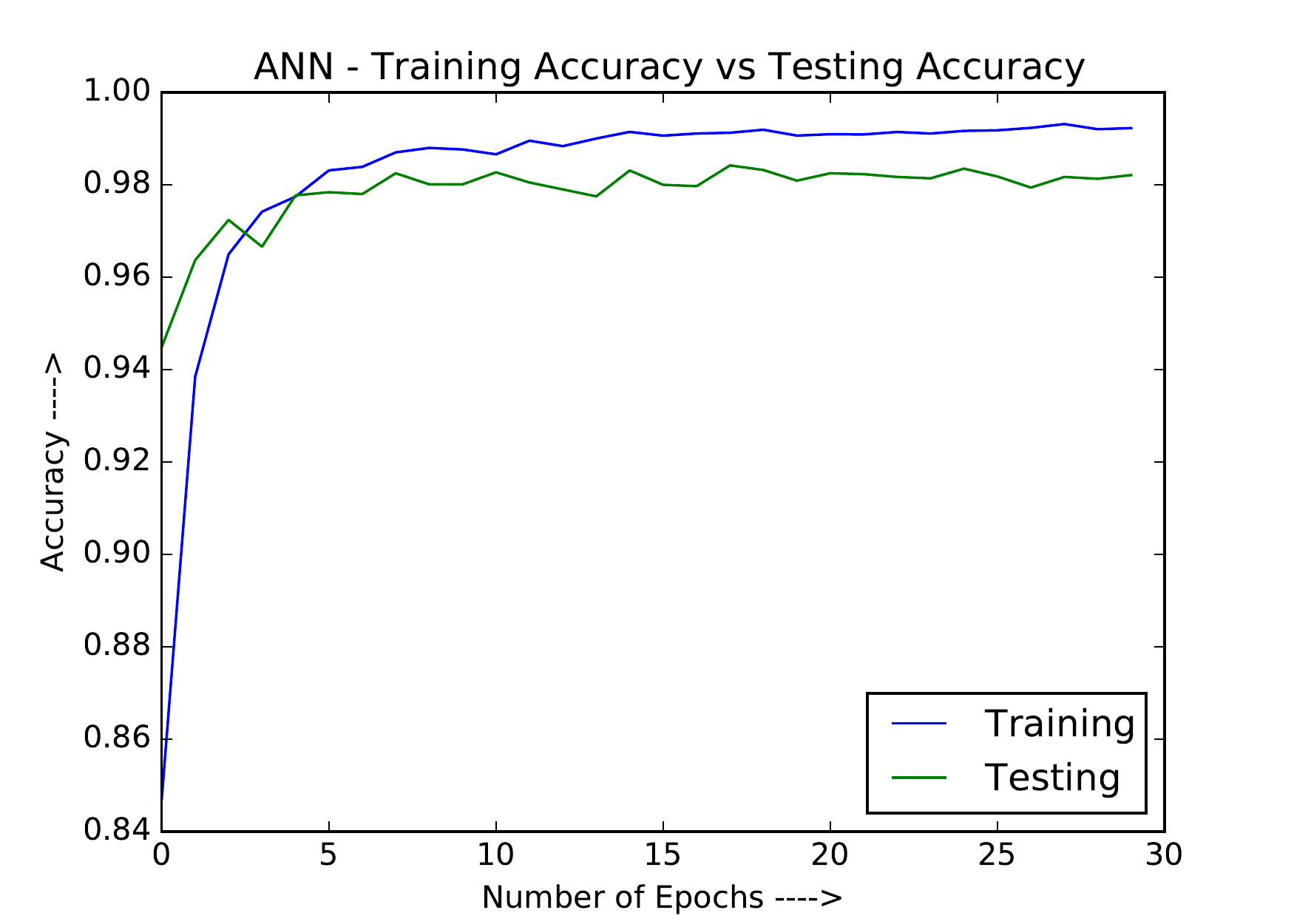}
\end{center}
\footnotesize Figure 12. Graph illustrating the transition of training accuracy with increasing number of epochs in Multilayer Perceptron (Accuracy v/s Number of epochs).

\begin{flushleft}
\includegraphics[scale=0.33]{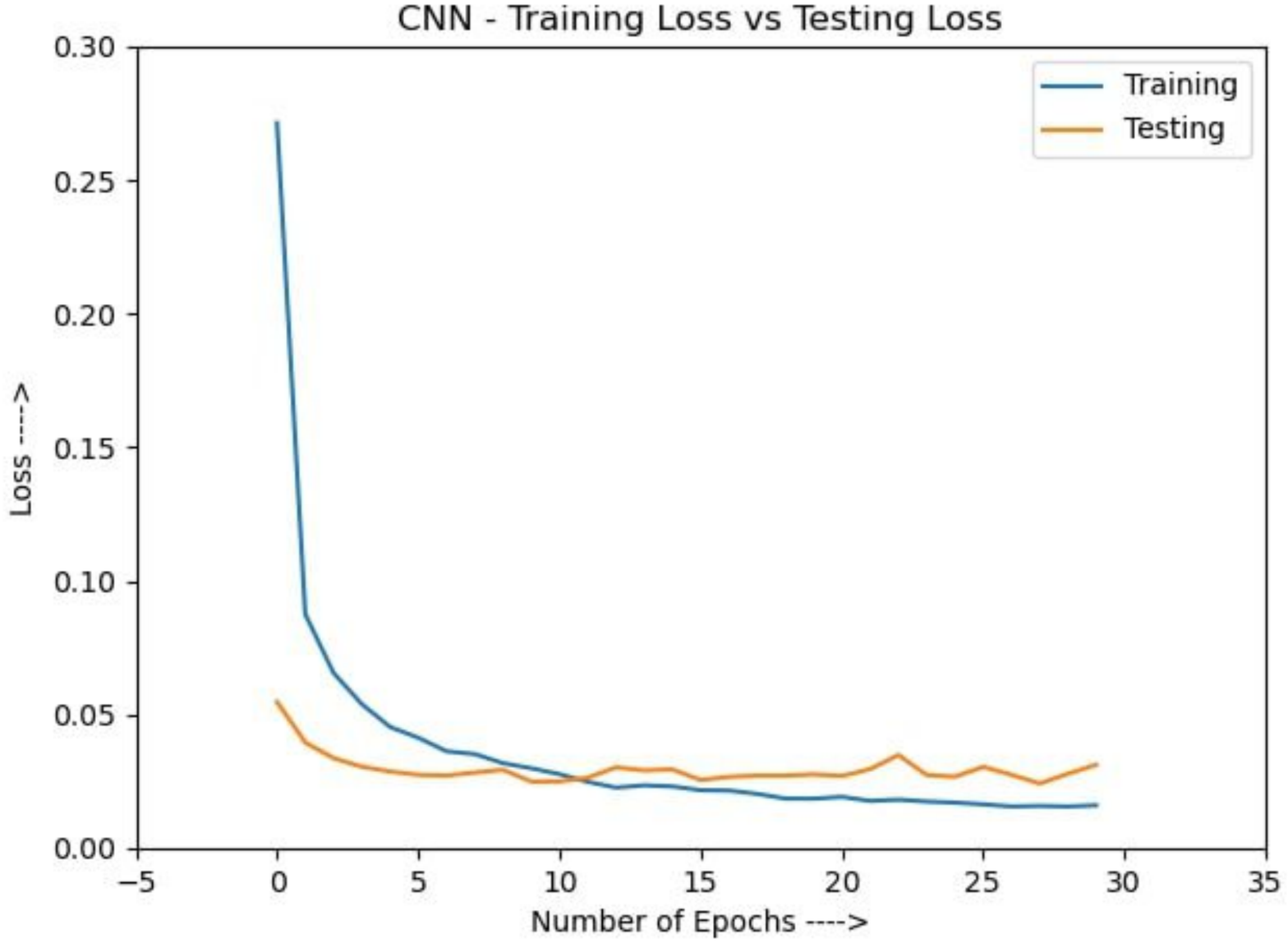}
\end{flushleft}
\footnotesize Figure 13. Graph illustrating the transition of training loss of CNN with increasing number of epochs (Loss rate v/s Number of epochs).

\begin{flushleft}
\includegraphics[scale=0.53]{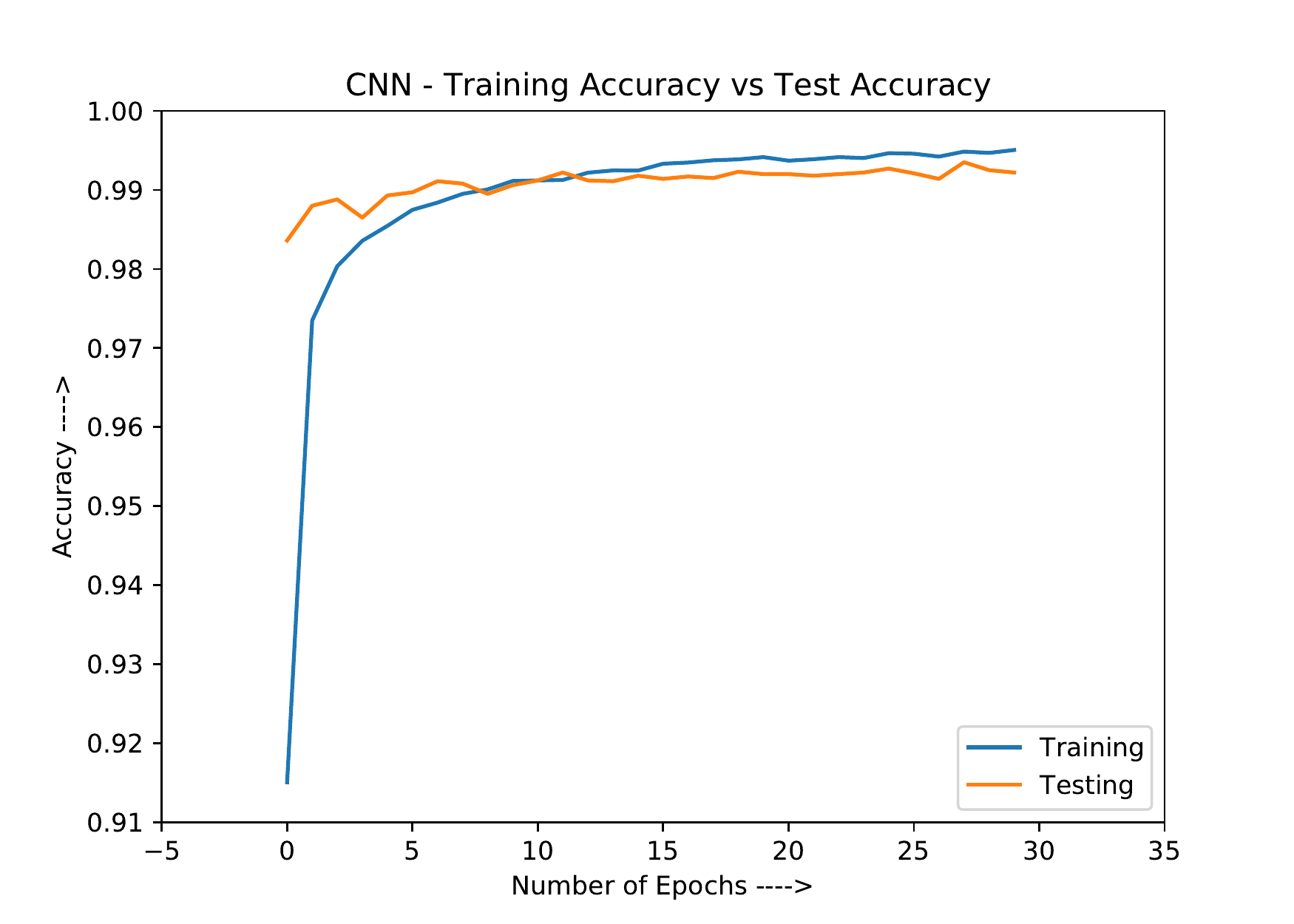}
\end{flushleft}
\footnotesize Figure 14. Graph illustrating the transition of training accuracy of CNN with increasing number of epochs (Accuracy v/s Number of epochs).
\end{small}

\section{\textbf{CONCLUSION}}
\begin{small}
In this research, we have implemented three models for handwritten digit recognition using MNIST datasets, based on deep and machine learning algorithms. We compared them based on their characteristics to appraise the most accurate model among them. Support vector machines are one of the basic classifiers that's why it’s faster than most algorithms and in this case, gives the maximum training accuracy rate but due to its simplicity, it’s not possible to classify complex and ambiguous images as accurately as achieved with MLP and CNN algorithms. We have found that CNN gave the most accurate results for handwritten digit recognition. So, this makes us conclude that CNN is best suitable for any type of prediction problem including image data as an input. Next, by comparing execution time of the algorithms we have concluded that increasing the number of epochs without changing the configuration of the algorithm is useless because of the limitation of a certain model and we have noticed that after a certain number of epochs the model starts overfitting the dataset and give us the biased prediction. \\

\end{small}

\section{\textbf{FUTURE ENHANCEMENT}}
\begin{small}
The future development of the applications based on algorithms of deep and machine learning is practically boundless. In the future, we can work on a denser or hybrid algorithm than the current set of algorithms with more manifold data to achieve the solutions to many problems. \\

In future, the application of these algorithms lies from the public to high-level authorities, as from the differentiation of the algorithms above and with future development we can attain high-level functioning applications which can be used in the classified or government agencies as well as for the common people, we can use these algorithms in hospitals application for detailed medical diagnosis, treatment and monitoring the patients, we can use it in surveillances system to keep tracks of the suspicious activity under the system, in fingerprint and retinal scanners, database filtering applications, Equipment checking for national forces and many more problems of both major and minor category. The advancement in this field can help us create an environment of safety, awareness and comfort by using these algorithms in day to day application and high-level application (i.e. Corporate level or Government level). Application-based on artificial intelligence and deep learning is the future of the technological world because of their absolute accuracy and advantages over many major problems. \\
\end{small}

\end{document}